\documentclass[lettersize,journal]{IEEEtran}
\usepackage{amsmath,amssymb,amsfonts}
\usepackage{algorithmic}
\usepackage{algorithm}
\usepackage{array}
\usepackage[caption=false,font=normalsize,labelfont=sf,textfont=sf]{subfig}
\usepackage{textcomp}
\usepackage{stfloats}
\usepackage{url}
\usepackage{verbatim}
\usepackage{graphicx}
\usepackage{cite}
\usepackage{wrapfig}
\usepackage{hyperref}
\usepackage{booktabs}
\usepackage{multirow} 

\begin{document}

\title{SA-MLP: A Low-Power Multiplication-Free Deep Network for 3D Point Cloud Classification in Resource-Constrained Environments}

\author{Qiang Zheng, Chao Zhang, and Jian Sun
}



\maketitle

\begin{abstract}
Point cloud classification plays a crucial role in the processing and analysis of data from 3D sensors such as LiDAR, which are commonly used in applications like autonomous vehicles, robotics, and environmental monitoring. However, traditional neural networks, which rely heavily on multiplication operations, often face challenges in terms of high computational costs and energy consumption. This study presents a novel family of efficient MLP-based architectures designed to improve the computational efficiency of point cloud classification tasks in sensor systems. The baseline model, Mul-MLP, utilizes conventional multiplication operations, while Add-MLP and Shift-MLP replace multiplications with addition and shift operations, respectively. These replacements leverage more sensor-friendly operations that can significantly reduce computational overhead, making them particularly suitable for resource-constrained sensor platforms. To further enhance performance, we propose SA-MLP, a hybrid architecture that alternates between shift and adder layers, preserving the network depth while optimizing computational efficiency. Unlike previous approaches such as ShiftAddNet, which increase the layer count and limit representational capacity by freezing shift weights, SA-MLP fully exploits the complementary advantages of shift and adder layers by employing distinct learning rates and optimizers. Experimental results show that Add-MLP and Shift-MLP achieve competitive performance compared to Mul-MLP, while SA-MLP surpasses the baseline, delivering results comparable to state-of-the-art MLP models in terms of both classification accuracy and computational efficiency. This work offers a promising, energy-efficient solution for sensor-driven applications requiring real-time point cloud classification, particularly in environments with limited computational resources.
\end{abstract}


\section{Introduction}
The analysis of point cloud data has become a cornerstone in various sensor-driven technological domains, such as autonomous driving~\cite{2020SalsaNet, 2021ADReview}, robotics~\cite{2020robot, 2020Gesture}, and virtual reality~\cite{2016Integrating, 2023paired}. Point clouds, which represent 3D spatial information through a collection of discrete points, are typically captured by LiDAR and other 3D sensors, providing rich data that is crucial for many applications. However, the inherently irregular and sparse nature of point clouds makes them challenging to process using traditional deep learning architectures originally designed for structured data like 2D images. As sensor systems become more prevalent in edge applications, real-time data processing is becoming increasingly important. This study focuses on improving the computational efficiency of point cloud classification algorithms for resource-constrained sensor platforms, which are often limited in terms of processing power and energy.

Over the years, advanced neural network architectures, such as multi-layer perceptrons (MLPs), convolutional neural networks (CNNs), graph neural networks (GNNs), and Transformers, have demonstrated significant potential in processing point cloud data~\cite{2017PointNet, 2018PointCNN, 2021PAConv, 2021point-trans}. Despite their effectiveness, these architectures often rely heavily on multiplication operations, which are computationally intensive and energy-demanding. Such reliance presents critical obstacles for their adoption in real-time sensor applications, particularly in edge devices or embedded systems where computational resources are constrained. This inefficiency is a major barrier in applications like autonomous driving and robotics, where sensor data processing must occur in real-time with limited energy consumption.

The computational inefficiency of multiplication operations in neural networks stems from their high energy consumption and increased clock-cycle demands compared to simpler alternatives such as addition or bitwise shifting. These inefficiencies become particularly problematic when deploying models on edge devices that are integral parts of sensor systems. This has motivated recent research into operation-efficient neural networks that replace multiplication with computationally lighter operations. For instance, AdderNet~\cite{2020AdderNet} replaces multiplications with addition operations in CNNs, achieving significant reductions in computational cost while maintaining competitive performance. Similarly, DeepShift~\cite{2021DeepShift} introduces bitwise shift operations as substitutes for multiplication, leveraging the simplicity of shift operations to achieve enhanced efficiency. These approaches have demonstrated that it is possible to maintain the performance of neural networks while significantly reducing their computational demands, thus making them more suitable for sensor-based applications. However, architectures like ShiftAddNet~\cite{2020ShiftAddNet}, which combine shift and addition operations, introduce their own challenges. Specifically, ShiftAddNet replaces convolutional layers with adder and shift layers, doubling both the parameter count and the number of layers. Additionally, it freezes the weights of shift layers, limiting the representational capacity and learning potential of the model. These trade-offs underscore the need for sensor-optimized approaches that balance computational efficiency with learning flexibility and representational power.

While much of the existing work on operation-efficient neural networks has focused on CNNs for image processing, the principles of replacing multiplication with addition and shift operations can be extended to MLP-based architectures. MLPs are widely used in point cloud processing due to their ability to capture point-wise features and their simplicity compared to CNNs. However, the reliance on multiplication-heavy layers makes MLPs computationally expensive and less suitable for real-time sensor processing. Unlike CNNs, which primarily rely on convolutional operations to process structured data, MLPs are designed to handle unstructured data such as point clouds through point-wise feature extraction. However, the absence of convolutional mechanisms in MLPs necessitates careful adaptation of multiplication-free designs to ensure compatibility with the unique requirements of point cloud data. Consequently, there is a need for novel strategies to achieve computational efficiency in MLPs while maintaining high classification performance, particularly in the context of sensor data processing for resource-constrained systems.

In this study, we propose a series of MLP-based architectures that address the computational inefficiencies of traditional point cloud classification models, with a focus on making them suitable for sensor applications. The baseline model, Mul-MLP, adopts the conventional MLP design and serves as a benchmark for comparison. To reduce computational costs, we introduce Add-MLP and Shift-MLP, which replace the multiplication operations in Mul-MLP with addition and bitwise shift operations, respectively. Add-MLP leverages addition operations based on the $L_{1}$ norm, achieving a significant reduction in computational cost while maintaining performance comparable to traditional MLPs. However, the expressiveness of Add-MLP is limited, as addition operations may struggle to capture certain multiplicative transformations. Similarly, Shift-MLP enhances computational efficiency by replacing multiplications with bitwise shifts but is inherently constrained by the scaling limitations of shift operations, which are restricted to powers of two. As a result, models relying solely on either addition or shift operations may lack the flexibility needed for complex tasks.

To address these limitations, we propose SA-MLP, a hybrid architecture that combines the strengths of addition and shift operations. SA-MLP replaces multiplication-heavy MLP layers with alternately distributed adder and shift layers, preserving the original depth and parameter count of the network. Unlike previous models such as ShiftAddNet~\cite{2020ShiftAddNet}, SA-MLP allows all parameters, including those in shift layers, to learn during training, avoiding the constraints imposed by frozen weights. By integrating addition and shift operations, SA-MLP leverages the complementary strengths of these approaches: the adder layers provide fine-grained flexibility for feature manipulation, while the shift layers enhance computational efficiency. This balanced design ensures that the model achieves high representational capacity without incurring unnecessary computational overhead.

A key innovation of SA-MLP is its tailored optimization strategy, which assigns distinct learning rates and optimizers to the adder and shift layers. This approach allows each type of operation to contribute optimally to the learning process, maximizing the complementary advantages of the two operations. By decoupling the optimization strategies, SA-MLP effectively combines computational efficiency with high accuracy, making it well-suited for sensor systems operating in real-time, where both energy consumption and performance are critical considerations.

Extensive experiments validate the effectiveness of the proposed architectures. Both Add-MLP and Shift-MLP achieve competitive performance compared to traditional MLPs while improving computational efficiency by leveraging addition and shift operations in place of multiplications. SA-MLP, in particular, surpasses the performance of the baseline Mul-MLP model and achieves results comparable to state-of-the-art MLP-based architectures. Additionally, the hybrid architecture offers a scalable and efficient solution for point cloud classification, providing a practical alternative for applications where computational efficiency and low energy consumption are paramount. This is especially true for sensor systems where resources such as power and processing capacity are limited, making it ideal for real-time deployment on edge devices like LiDAR-based systems in autonomous vehicles and robotic systems.

This study makes the following contributions: 
\begin{itemize} 
\item We introduce Add-MLP and Shift-MLP, novel MLP-based architectures that replace multiplication operations with addition and shift operations, respectively, improving computational efficiency in point cloud classification tasks. 
\item We propose SA-MLP, a hybrid architecture that combines adder and shift layers within MLPs, maintaining the original network depth and parameter count while allowing all parameters to learn during training. 
\item We develop a tailored optimization strategy with distinct learning rates and optimizers for adder and shift layers, effectively leveraging their complementary strengths. 
\item Experimental evaluations demonstrate that SA-MLP outperforms the traditional multiplication-based Mul-MLP model and achieves performance comparable to state-of-the-art MLP-based models, offering a more efficient and scalable solution for point cloud classification.
\end{itemize}

\section{Related Works} 
The field of 3D point cloud analysis has experienced significant advancements, particularly with the integration of deep learning techniques that effectively process unstructured data for tasks such as object recognition and scene understanding. Concurrently, the pursuit of computational efficiency has led to the development of multiplication-free neural networks, which aim to reduce computational complexity by substituting traditional multiplication operations with more efficient alternatives. The following subsections delve into these two pivotal areas, providing an overview of key methodologies and their contributions to the advancement of efficient 3D point cloud analysis.

\subsection{Point Cloud Analysis}
The analysis of 3D point clouds has become a focal point in computer vision, especially for applications in sensor-based systems such as object recognition, scene understanding, and autonomous navigation. Point clouds, which are typically generated by sensors like LiDAR (Light Detection and Ranging), depth cameras, and stereo vision systems, provide rich spatial data that can capture the geometry of physical environments. These sensors are foundational in fields such as autonomous vehicles, robotics, and geospatial mapping. Early breakthroughs in point cloud processing leveraged deep learning techniques to handle the unstructured nature of point cloud data, marking a departure from traditional grid-based representations used in 2D image processing.

One of the seminal works in this area is PointNet~\cite{2017PointNet}, which demonstrated that deep neural networks could directly process point clouds without requiring structured grid-like representations, a significant departure from traditional methods. PointNet++~\cite{2017PointNetplus} extended this idea by introducing a hierarchical architecture that captures local and global geometric information, thus improving the model's ability to learn from complex spatial structures.

Alongside these pioneering approaches, Convolutional Neural Networks (CNNs) have been adapted to process point cloud data. One notable example is PointCNN~\cite{2018PointCNN}, which generalizes the convolution operation to point clouds through the introduction of the $\mathcal{X}$-Transformation, effectively transforming point clouds into a more structured form suitable for convolution. Another promising direction involves the exploration of continuous convolutional networks~\cite{2021Deep}, which aim to learn parametric filters for point cloud data, enabling more flexible and efficient feature extraction.

Graph Neural Networks (GNNs) have also shown great potential in point cloud analysis. Dynamic Graph CNN (DGCNN)~\cite{2019DGCNN} introduced a dynamic graph construction method, which adapts to the local geometry of the point cloud, allowing the network to focus on relevant local structures. Building on this, Adaptive Graph Convolution~\cite{2021PAConv} proposed an adaptive graph convolution scheme to further refine the feature extraction process by considering the varying local point cloud characteristics. 

In contrast to these graph-based methods, Transformer-based architectures have gained traction in point cloud analysis, with Point Transformer~\cite{2021point-trans} being a notable example. Utilizing the self-attention mechanism, Point Transformer models capture global dependencies between points, enabling more robust feature learning. Point Transformer V2~\cite{2022point-transV2} extended this idea by incorporating grouped vector attention and partition-based pooling, which further enhances the model's ability to model complex dependencies and structures in point clouds.

In summary, the integration of deep learning techniques with sensor-based point cloud data has led to significant advancements in the analysis of 3D data. However, while these methods have shown promising results in research settings, challenges remain in terms of real-time implementation and computational efficiency, especially in the context of embedded sensor systems and resource-constrained environments. The evolution of models like PointNet~\cite{2017PointNet}, PointCNN~\cite{2018PointCNN}, and Point Transformer~\cite{2021point-trans} reflects the growing sophistication of algorithms tailored to the unique characteristics of sensor data, with the potential to significantly impact fields such as autonomous navigation, robotics, and environmental sensing.

\subsection{Multiplication-Free Neural Networks}
In parallel with the advancements in point cloud analysis, the increasing need for computational efficiency in sensor-based systems has driven the exploration of multiplication-free neural networks. These networks aim to reduce the computational burden typically associated with traditional multiplication operations, which are computationally expensive, by replacing them with simpler, faster, and energy-efficient operations. This is particularly crucial for embedded sensor systems, such as autonomous vehicles, drones, and IoT devices, where real-time data processing, low latency, and low power consumption are essential.

Early efforts in this direction focused on binarizing weights and activations, as seen in Binary Neural Networks (BNN)\cite{2016BNN} and XNOR-Net\cite{2016XNOR}. These approaches replace floating-point multiplications with binary operations, significantly reducing both computational load and memory footprint. These approaches replace floating-point multiplications with binary operations, significantly reducing both the computational load and memory footprint, thus making them suitable for deployment in resource-constrained environments.

Shift-based operations also offer a promising alternative to multiplication. For instance, methods like SSL~\cite{2019SSL}, Shift~\cite{2018Shift}, and DeepShift~\cite{2021DeepShift} leverage bitwise shifts, which are highly efficient on hardware, to replace multiplication. These approaches capitalize on the efficiency of simple shift operations to accelerate neural network implementations. Another key advancement is the use of additive operations in networks, such as AdderNet~\cite{2020AdderNet}, which replaces multiplications with additions. The AdderNet framework, along with subsequent works like AdderSR~\cite{2021AdderSR} and AdderFPGA~\cite{2021AdderFPGA}, demonstrates that high performance can still be achieved with additive operations, challenging the traditional reliance on multiplication.

A more recent trend combines both shift and add operations in hybrid models. ShiftAddNet~\cite{2020ShiftAddNet} and ShiftAddViT~\cite{2023ShiftAddViT} are examples of architectures that integrate both operations, providing a flexible trade-off between computational efficiency and model accuracy. These hybrid models exploit the advantages of both shift and add operations, making them particularly effective in reducing the computational overhead while maintaining competitive performance.

Despite the significant advancements in multiplication-free architectures, their application to point cloud analysis remains relatively underexplored. Most existing work on multiplication-free networks has focused on tasks like image classification, which do not have the inherent complexities of 3D point cloud data. Point cloud analysis—especially in sensor-based systems that rely on LiDAR or depth sensors—presents unique challenges due to the high dimensionality and unstructured nature of the data. This study contributes to filling this gap by introducing shift and add operations for 3D point cloud classification. Through extensive experimentation, we demonstrate that these multiplication-free architectures not only reduce computational costs but also achieve competitive classification accuracy when compared to traditional models, thereby providing a promising solution for efficient point cloud analysis in resource-constrained environments.

\section{Methodology}
This section outlines the methodologies underpinning our efficient point cloud classification models. We begin with the conventional Multiplication MLP (Mul-MLP) as a baseline, then introduce the Add-MLP and Shift-MLP models, which respectively replace multiplications with additions and bitwise shifts to enhance computational efficiency. Advancing further, the ShiftAdd-MLP (SA-MLP) model harmoniously combines these operations, aiming to achieve a synergistic balance between efficiency and representational capacity. The section concludes with a discussion on the optimization strategies tailored for the proposed models, emphasizing the assignment of distinct learning rates and optimizers to different layer types to fully harness their individual advantages in enhancing classification accuracy and efficiency.

\subsection{Baseline Model: Mul-MLP}
\begin{figure}[htbp]
    \centering
    \includegraphics[width=\linewidth]{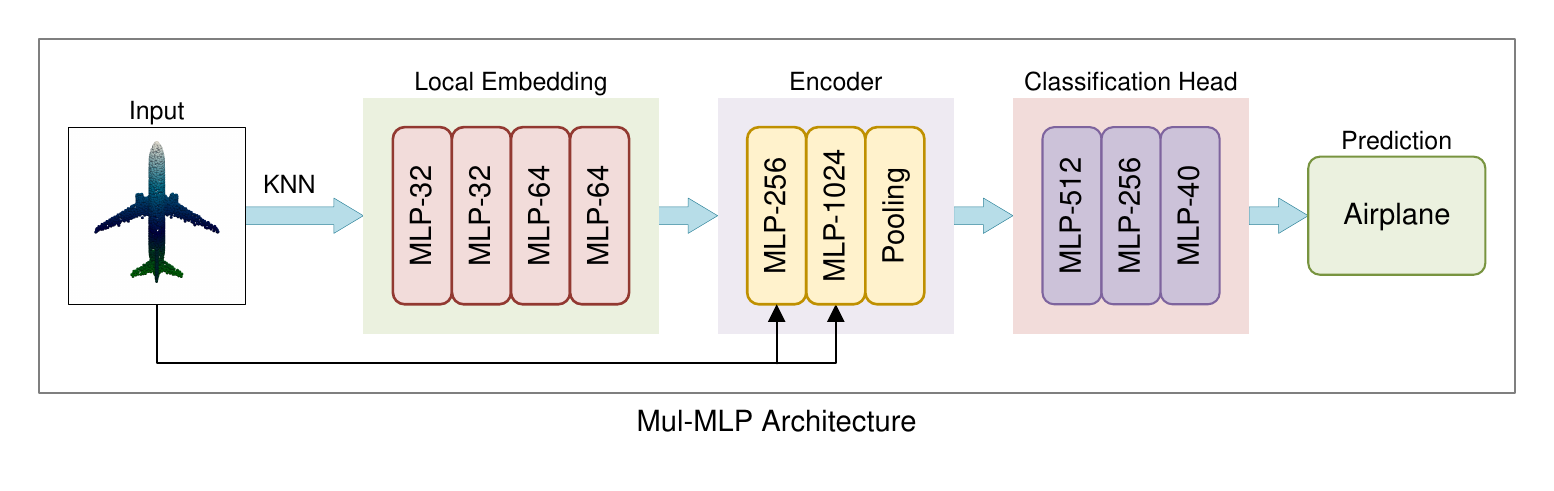}
    \caption{Architecture of the Multiplication-Based MLP (Mul-MLP) Model. This diagram illustrates the structure of the Mul-MLP model, where all layers utilize traditional multiplication operations for feature extraction and classification.}
    \label{fig-SA-MLP-mul}
\end{figure}

Inspired by the use of shared MLPs in MLP-based models~\cite{2017PointNet, 2022Fusion}, and the tokenization approach for obtaining local embeddings and complementary positional information in Vision Transformers (ViT)~\cite{2021ViT}, this study proposes a baseline model named Multiplication MLP (Mul-MLP) for point cloud classification. The Mul-MLP model is designed with a concise structure that effectively captures the essential features of 3D point clouds while maintaining computational efficiency, illustrated in Fig~\ref{fig-SA-MLP-mul}.

The Mul-MLP model consists of three primary components: a local embedding module, an encoder, and a classification header. Given an input point cloud \( P \in \mathbb{R}^3 \), the local embedding module first extracts local embedding features for each point. This is achieved by considering the local geometric context of each point, thereby capturing both the spatial relationships and the intrinsic properties of the point cloud. The extracted local embedding features are then passed into the encoder, which forms the core of the Mul-MLP model. Each layer of the encoder concatenates the input features with the corresponding global point coordinates before processing them. By integrating the local features with their spatial positions at every layer, the encoder ensures that both the detailed local geometry and the broader spatial distribution of points are effectively captured, allowing the model to maintain an awareness of the spatial arrangement throughout the network. The fusion of local and global information at each layer enhances the model's ability to learn complex spatial patterns. After passing through the encoder, a global feature representation of the shape is obtained via maximum pooling across all points. This pooled feature, which represents the overall structure of the point cloud, is then fed into the classification header. The classification header consists of a series of fully connected layers that ultimately output the class prediction for the input point cloud.

The Mul-MLP model serves as a foundational baseline for this study, providing a benchmark against which the proposed variants, Add-MLP, Shift-MLP, and SA-MLP, are compared. By leveraging the simplicity and efficiency of MLPs, Mul-MLP sets a strong baseline in terms of both performance and computational cost for point cloud classification tasks.

\subsection{Shift-Based Model: Shift-MLP}
The Shift-MLP model proposed in this study, inspired by DeepShift~\cite{2021DeepShift}, introduces an innovative approach by replacing the traditional multiplication operations in MLP layers with bitwise shift operations, significantly improving computational efficiency without compromising model performance. The underlying principle of the shift layer involves quantizing the weights to powers of two, thereby transforming multiplication into efficient shift operations. In the forward propagation, each weight \( W \) is first quantized using the formulas \( S = \text{sign}(W) \) and \( p = \text{round}(\log_2(|W|)) \), resulting in a quantized weight:

\begin{equation}
W_s = S \cdot 2^p
\end{equation}

This quantization allows the model to compute the output as:

\begin{equation}
Y = W_s \cdot X
\end{equation}

\begin{figure}[htbp]
    \centering
    \includegraphics[width=\linewidth]{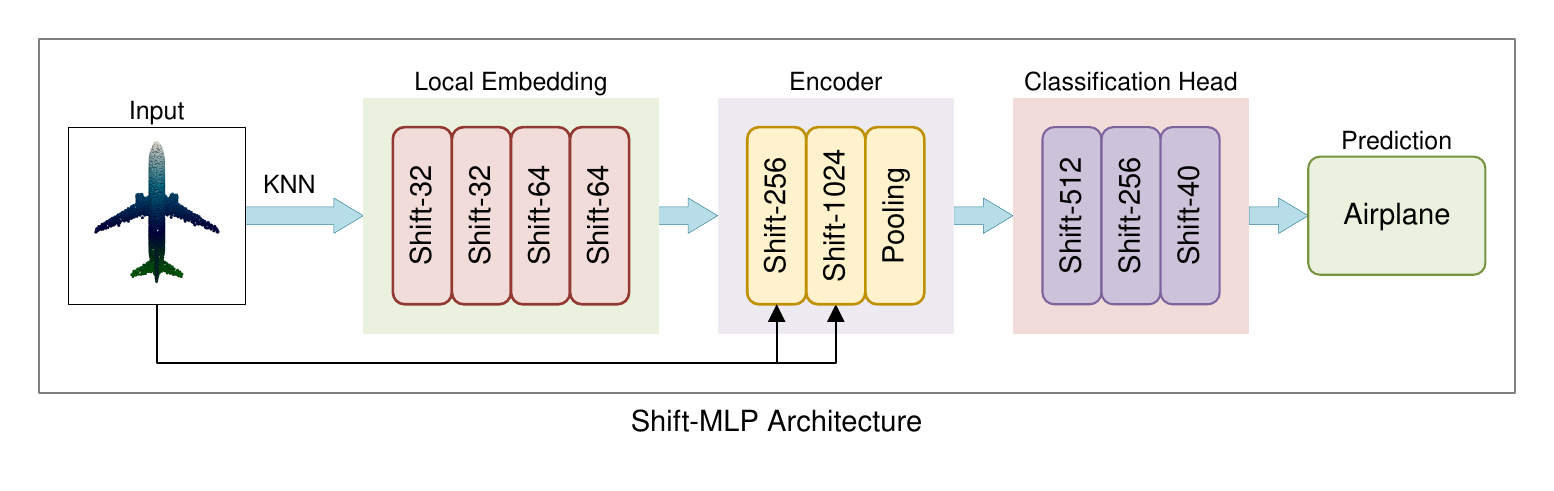}
    \caption{Architecture of the Shift-Based MLP (Shift-MLP) Model. This diagram shows the structure of the Shift-MLP model, where all layers employ bitwise shift operations to achieve computational efficiency.}
    \label{fig-SA-MLP-shift}
\end{figure}

This approach effectively reduces the computational cost associated with floating-point multiplications. The network structure diagram for Shift-MLP is shown in Figure \ref{fig-SA-MLP-shift}.

In the backpropagation process, the gradients are computed similarly to traditional MLPs, with minor adjustments due to the quantization step. The partial derivative of the loss \( L \) with respect to the input \( X \) is given by:

\begin{equation}
\frac{\partial L}{\partial X} = \frac{\partial L}{\partial Y} \cdot W_s
\end{equation}
while the gradient with respect to the weight \( W \) is approximated as:

\begin{equation}
\frac{\partial L}{\partial W} \approx \frac{\partial L}{\partial Y} \cdot X
\end{equation}
assuming \( \frac{\partial W_s}{\partial W} \approx 1 \). This assumption simplifies the backpropagation, allowing the model to leverage the efficient forward computation without altering the training dynamics significantly. It’s important to note that during training, the quantization is only applied during the forward pass, leaving the backpropagation process consistent with traditional MLPs. This design choice ensures that the model benefits from the efficiency of shift operations while maintaining the learning capabilities of standard MLPs.

The Shift-MLP network, built using the described shift layers, demonstrates a balanced trade-off between computational efficiency and model accuracy. The model operates by converting the input into a 32-bit fixed-point representation, with 16 bits allocated for the integer part and 16 bits for the fractional part. The weights are stored in a compact 5-bit format, where 1 bit represents the sign \( S \), and the remaining 4 bits store the shift amount \( p \). To optimize the quantization process, the weights \( W \) are clamped to lie within the range of ±1 before quantization, based on statistical analyses of traditional MLP networks, which have shown that most weights fall within this range after training. This clamping ensures that only left shifts are required, allowing the 4 bits to be used exclusively for storing the shift amount \( p \), without needing to save the sign of the shift. By confining the weights within a small range close to zero, the quantization error is minimized, thereby maintaining high accuracy despite the reduced precision. This design consideration significantly enhances the model's performance while preserving computational efficiency, making Shift-MLP a powerful approach for point cloud classification.

\subsection{Addition-Based Model: Add-MLP}
\begin{figure}[htbp]
    \centering
    \includegraphics[width=\linewidth]{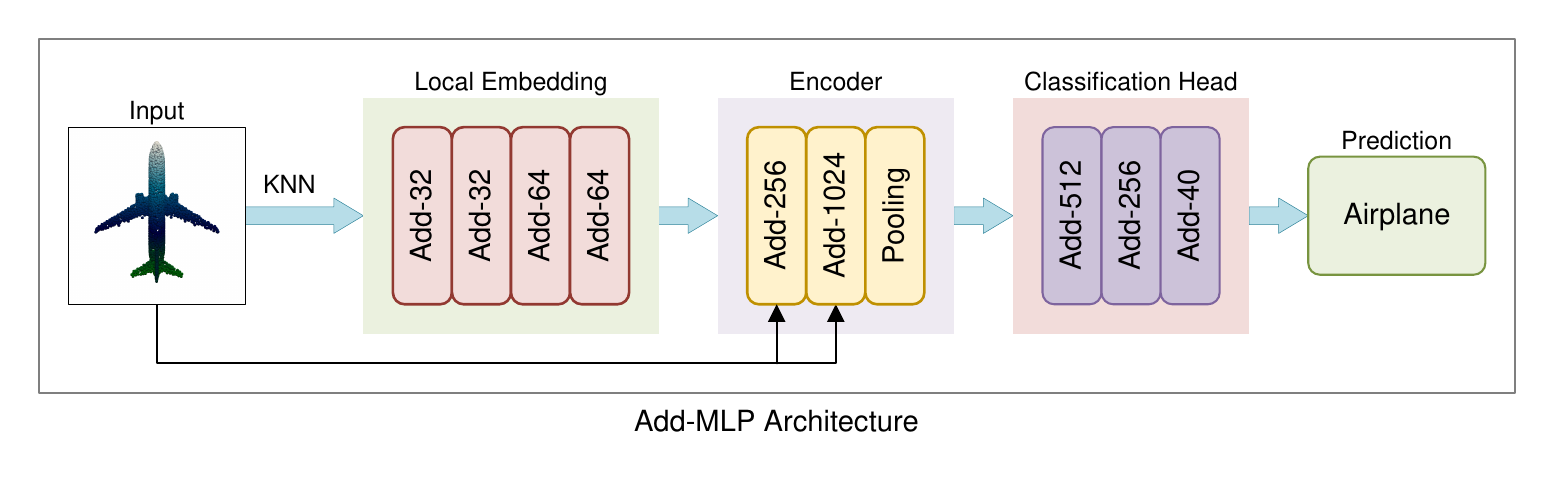}
    \caption{Architecture of the Addition-Based MLP (Add-MLP) Model. This diagram presents the structure of the Add-MLP model, which replaces traditional multiplication operations with adder layers throughout the network.}
    \label{fig-SA-MLP-add}
\end{figure}

In this study, we propose Add-MLP, a MLP-based model designed for point cloud classification, as illustrated in Fig.~\ref{fig-SA-MLP-add}. The Add-MLP model is inspired by the design principles of AdderNet~\cite{2020AdderNet}, replacing traditional multiplication-based operations with addition to achieve greater computational efficiency without sacrificing model accuracy. Unlike conventional models that rely on dot products to compute activations, Add-MLP leverages the $L_1$ norm between the input features \( X \) and the weights \( W \) as a measure of relevance. This substitution provides a significant computational advantage, particularly in resource-constrained environments.

The forward propagation process in Add-MLP is governed by the following equation:
\begin{equation}
Y = - \parallel X-W \parallel_1
\end{equation}
where \( Y \) is the output, \( X \) is the input, and \( W \) represents the corresponding weights. The use of the $L_1$ norm allows the network to measure relevance based on additive differences, which fundamentally reduces the computational costs compared to traditional multiplication-based operations.

The backward propagation process in Add-MLP introduces key adjustments to the gradient calculation. Initially, the gradient of the loss \( L \) with respect to the weights \( W \) is derived as:
\begin{equation}
\frac{\partial L}{\partial W} = \frac{\partial L}{\partial Y} \cdot \text{sign}(X - W)
\end{equation}
This formulation reflects the nature of the $L_1$ norm, where the gradient is typically defined by the sign of the difference between \( X \) and \( W \). However, in practice, this leads to challenges during training due to the discrete nature of the sign function, which can cause unstable weight updates and hinder convergence. To address this, the gradient is modified to:
\begin{equation}
\frac{\partial L}{\partial W} = \frac{\partial L}{\partial Y} \cdot (X - W)
\end{equation}
This modification smooths the gradient, converting the coarse sign-based updates into more granular adjustments. The transition from \(\text{sign}(X - W)\) to \(X - W\) ensures that the gradient retains continuous information, facilitating more stable and effective learning. 

Similarly, the gradient of the loss with respect to the input \( X \) is initially calculated as:
\begin{equation}
\frac{\partial L}{\partial X} = -\frac{\partial L}{\partial Y} \cdot \text{sign}(X - W)
\end{equation}
and then modified to:
\begin{equation}
\frac{\partial L}{\partial X} = -\frac{\partial L}{\partial Y} \cdot (X - W)
\end{equation}
This modification, like with the gradient for \( W \), ensures a smooth and continuous gradient, which is essential for stable training. However, unlike the gradient for \( W \), the gradient of \( X \) impacts all preceding layers due to the chain rule in backpropagation. This accumulation of gradients across multiple layers can lead to gradient explosion, causing instability in the model’s training. To mitigate this risk, a clipping function is applied:
\begin{equation}
\frac{\partial L}{\partial X} = -\frac{\partial L}{\partial Y} \cdot \text{clip}(X - W)
\end{equation}
where:
\begin{equation}
\text{clip}(x) = \begin{cases}
  1,    & \text{if } x > 1,             \\
  x,    & \text{if } -1 \leq x \leq 1,  \\
  -1,   & \text{if } x < -1.
\end{cases}
\end{equation}
The clipping operation ensures that the gradient remains within a bounded range, thereby stabilizing the training process and preventing the runaway growth of gradients that could lead to numerical instability.

The Add-MLP network incorporates these adder layers to achieve a balance between computational efficiency and model performance. By shifting from multiplication to addition and making thoughtful adjustments to the gradients, Add-MLP offers a robust and accurate approach to point cloud classification. These innovations ensure that the model remains stable during training, without sacrificing the quality of its predictions. Add-MLP exemplifies a novel approach to neural network design, showing that with thoughtful adjustments to core operations and gradient calculations, it is possible to achieve both high efficiency and robust performance. The adjustments to the gradients of both \( W \) and \( X \) are key to the network’s success, ensuring stability and effectiveness throughout the training process.

\subsection{Combined Model: ShiftAdd-MLP (SA-MLP)}
\begin{figure}[htbp]
    \centering
    \includegraphics[width=\linewidth]{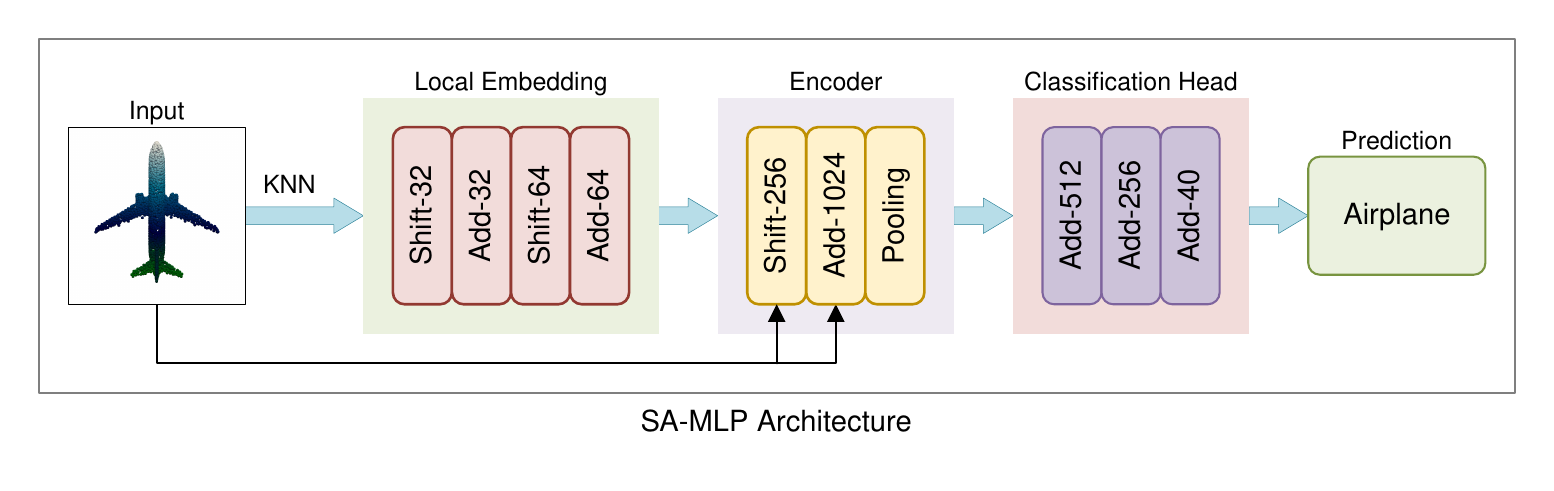}
    \caption{Architecture of the ShiftAdd-MLP (SA-MLP) Model. This diagram depicts the structure of the SA-MLP model, integrating both adder and shift layers to leverage the complementary strengths of each operation.}
    \label{fig-SA-MLP-sa}
\end{figure}

The SA-MLP model introduced in this study builds on the concepts of Shift-MLP and Add-MLP, offering an advanced approach that fully leverages the complementary strengths of shift and add operations (see Fig.~\ref{fig-SA-MLP-sa}). Unlike ShiftAddNet~\cite{2020ShiftAddNet}, which operates in the image domain and doubles the number of layers by replacing each convolutional layer with a pair of Shift and Adder layers, SA-MLP integrates these operations in an interleaved manner within the MLP framework. The key innovation lies in replacing the traditional MLP layers in the baseline Mul-MLP model with a sequence of interleaved Shift and Adder layers. This design ensures that the number of layers and parameters remains consistent with the original model, avoiding the significant increase in computational costs observed in ShiftAddNet~\cite{2020ShiftAddNet}.

A critical distinction between SA-MLP and ShiftAddNet~\cite{2020ShiftAddNet} lies in the treatment and optimization of the shift and adder layers. In ShiftAddNet~\cite{2020ShiftAddNet}, the shift layer is frozen during training, preventing it from contributing to learning and reducing it to a source of random perturbation, which can degrade overall network performance. In contrast, SA-MLP allows both shift and adder layers to actively participate in the learning process by using differentiated optimization strategies tailored to the characteristics of each layer type. Although these strategies are only briefly mentioned here, they are vital to SA-MLP’s superior performance and will be detailed in the following subsection (Sec.~\ref{sec-SA-MLP-optim}), which discusses the optimization techniques for Shift-MLP, Add-MLP, and SA-MLP.

The motivation for combining shift and add operations in SA-MLP stems from the unique yet complementary strengths of each. Shift operations are highly efficient, reducing computational overhead through bitwise shifts, while add operations enhance feature extraction by enabling richer non-linear transformations. By interleaving these operations, SA-MLP captures a wide range of features with minimal parameter overhead, achieving an optimal balance between computational cost and performance, which is particularly advantageous in resource-constrained environments.

The SA-MLP network structure demonstrates how interleaved shift and adder layers systematically replace the traditional MLP layers. This design maintains the depth and parameter count of the original Mul-MLP model while improving its ability to extract and process features from point cloud data. The careful integration of these operations within the network allows SA-MLP to achieve high computational efficiency and model accuracy, making it an effective solution for point cloud analysis tasks where both performance and resource efficiency are critical.

\subsection{Optimization Strategies for Shift and Adder Layers in SA-MLP}\label{sec-SA-MLP-optim}
To address the distinct characteristics of the shift and adder layers in the SA-MLP model, this subsection elaborates on the necessity of individualized optimization strategies for each type. The shift layer, being structurally similar to the traditional multiplication-based MLP, requires minimal adjustments in optimization. In contrast, the adder layer, which fundamentally diverges from conventional operations, demands a more nuanced approach to ensure effective parameter updates.

\subsubsection{Shift Layer Optimization}
The shift layer replaces traditional multiplication with shift operations, incorporating quantization of weights to powers of two. Despite this modification, the forward and backward propagation processes remain closely aligned with those of the traditional multiplication-based MLP model. Consequently, the gradients and weight distributions do not significantly deviate, allowing for the adoption of standard optimization strategies. This study applies a cyclical learning rate schedule, coupled with the Adam optimizer, which has proven effective for similar architectures.

\subsubsection{Adder Layer Optimization}
The adder layer, on the other hand, introduces a substantial deviation from conventional architectures. It computes the correlation between input \(X\) and weight \(W\) using the negative $L_1$ norm, \(- \parallel X - W \parallel_1\), which results in negative outputs that approach zero as the correlation strengthens. This contrasts sharply with traditional MLPs and CNNs, where stronger correlations yield higher, unbounded activation values. This fundamental difference leads to distinct output distributions and a significant divergence in gradient behavior between the adder and shift layers.

The gradient calculation for the adder layer incorporates the factor \((X - W)\), and at high activation values, this term approaches zero, causing a sharp decline in gradient magnitude. This can result in ineffective updates if the same optimization strategy as the shift layer is applied. Additionally, the consistent subtraction of \(X\) and \(W\) necessitates that their magnitudes are aligned. Given that input \(X\) often undergoes batch normalization, it is crucial to normalize the gradient of \(W\) to maintain effective learning.

To address these challenges, this study employs a gradient modulation strategy inspired by AdderNet~\cite{2020AdderNet}. The gradient modulation is represented by the following formula:
\begin{equation}
\tilde{g} = \frac{\eta \nabla L(W)}{\parallel \nabla L(W) \parallel_2 / \sqrt{n}}
\label{eq-SA-MLP-modulation}
\end{equation}
where \(\tilde{g}\) is the modulated gradient, \(\eta\) is a hyperparameter that adjusts the gradient magnitude, \(n\) is the number of elements in \(W\), and \(\nabla L(W)\) is the gradient of the loss \(L\) with respect to \(W\).

The update to weight \(W\) is then calculated as:
\begin{equation}
\Delta W = l \times \tilde{g}
\end{equation}
where \(\Delta W\) represents the update amount, and \(l\) is the learning rate. Additionally, because Adam’s adaptive gradient adjustments might conflict with the gradient modulation strategy, the SGD optimizer is chosen for the adder layer. This choice ensures compatibility with the gradient adjustment mechanism and improves update accuracy.

\subsubsection{Integrated Optimization in SA-MLP}
The distinction between the shift and adder layers necessitates carefully designed optimization strategies to ensure the overall performance of the SA-MLP model. While ShiftAddNet~\cite{2020ShiftAddNet} previously opted to freeze the shift layer to maintain optimization stability, this approach compromises the model's adaptability. By contrast, this study’s individualized optimization for each layer type allows the SA-MLP to fully leverage the complementary strengths of both the shift and adder operations. The integration of shift and add operations in SA-MLP combines the computational efficiency of shifts with the enhanced feature extraction capabilities of additions. This hybrid approach strikes a balance between computational cost and model performance, making SA-MLP particularly suitable for scenarios where both efficiency and accuracy are crucial.

\section{Experiments}
This section begins by detailing the parameter configurations, followed by a presentation of the experimental results for point cloud classification. Finally, to provide deeper insights into the design rationale behind the SA-MLP model, we include visualizations and statistical analyses. 

\subsection{Training Configuration}
To ensure a fair comparison across all models, this study utilizes a uniform framework and consistent training parameters. The evaluation is conducted on the point cloud classification task using the ModelNet40 benchmark, which comprises 12,311 samples across 40 categories, with 9,843 samples for training and 2,468 for testing. Each point cloud input contains 1,024 points. A periodic annealing learning rate is employed with a batch size of 32 for all models. For Mul-MLP and Shift-MLP, the learning rate starts at \(10^{-3}\) and is reduced to \(10^{-6}\), with the Adam optimizer used for training. Conversely, Add-MLP employs a higher initial learning rate of \(2 \times 10^{-2}\) that decreases to \(2 \times 10^{-3}\) and utilizes the SGD optimizer. The SA-MLP model integrates a hybrid optimization strategy: the shift layers follow the same parameters as Shift-MLP, while the adder layers adopt the approach used in Add-MLP. This setup allows each model to be trained under consistent conditions while accounting for their unique optimization requirements.

\subsection{Classification Task}
In this subsection, we evaluate the performance of the SA-MLP model proposed in this study against several state-of-the-art methods on the ModelNet40 classification task. The results of this comparative analysis are summarized in Tab. \ref{tab-SA-MLP-cls}.

\begin{table}[ht]
    \centering
    \newcommand{\tabincell}[2]{\begin{tabular}{@{}#1@{}}#2\end{tabular}}
    \small
    \setlength{\tabcolsep}{3.2mm}
    \begin{tabular}{l|cccc}
        \toprule[1pt]
        \textbf{Method}                                 & Input         & Num.  & Acc. (\%) \\
        \midrule[0.3pt]
        PointNet~\cite{2017PointNet}					& xyz           & 1k    & 89.2 \\
        PointNet++(MSG)~\cite{2017PointNetplus}			& xyz, nor      & 5k    & 91.9 \\
        PointCNN~\cite{2018PointCNN}					& xyz           & 1k    & 92.2 \\
        DGCNN~\cite{2019DGCNN}					        & xyz           & 1k    & 92.9 \\
        KPConv~\cite{2019KPConv}					    & xyz           & 6.8k  & 92.9 \\
        PointNext~\cite{2022PointNext}                  & xyz           & 1k    & 93.2 \\
        AdaptConv~\cite{2021Adaptive}			        & xyz           & 1k    & 93.4 \\
        PointMixer~\cite{2021pointmixer}                & xyz           & 1k    & 93.6 \\
        PT~\cite{2021point-trans}                       & xyz           & 1k    & 93.7 \\
        \midrule[0.3pt]
        Mul-MLP                                         & xyz           & 1k    & 93.5 \\
        Shift-MLP                                       & xyz           & 1k    & 93.3 \\
        Add-MLP                                         & xyz           & 1k    & 93.1 \\
        SA-MLP                                          & xyz           & 1k    & 93.9 \\
        \bottomrule[1pt]
    \end{tabular}
    \vspace{5pt}
    \caption{Results for the ModelNet40 classification task.}
    \label{tab-SA-MLP-cls}
\end{table}

Tab. \ref{tab-SA-MLP-cls} illustrates the performance of various MLP-based methods, including PointNet~\cite{2017PointNet}, PointNet++\cite{2017PointNetplus}, and PointNext\cite{2022PointNext}. Remarkably, the SA-MLP model proposed in this study surpasses these established methods, achieving a classification accuracy of 93.9\%. This notable improvement underscores the efficacy of combining both shift and add operations within the MLP framework, as demonstrated in this research. The inclusion of these operations enhances the model's ability to extract and process point cloud features, contributing to its superior performance.

In addition to benchmarking SA-MLP against external methods, the study also evaluates the performance of four models developed as part of this research: Mul-MLP, Shift-MLP, Add-MLP, and SA-MLP. Mul-MLP serves as the baseline model, achieving an accuracy of 93.5\%, which establishes a strong foundation for comparison. Shift-MLP and Add-MLP, which replace multiplication operations with shift and add operations, respectively, achieve slightly lower accuracies of 93.3\% and 93.1\%. While both models demonstrate competitive performance and offer substantial improvements in computational efficiency, their slight reduction in accuracy reflects the trade-off typically encountered when replacing multiplication operations with more efficient alternatives.

However, the hybrid SA-MLP model, which synergistically integrates both shift and add operations, significantly outperforms the baseline, achieving an accuracy of 93.9\%. This result demonstrates the complementary strengths of the two operations—shift operations contribute to efficient computation, while add operations enhance the model's ability to capture complex feature representations. The interleaving of these two operations enables SA-MLP to strike a balance between efficiency and accuracy, making it particularly suitable for resource-constrained environments where computational cost is a critical factor.

This comparative analysis highlights the promising potential of developing multiplication-free architectures for point cloud analysis. While both Shift-MLP and Add-MLP deliver increased efficiency, it is the hybrid design of SA-MLP that capitalizes on their combined advantages, resulting in a significant performance gain. The results of this study pave the way for future research into the development of more efficient and effective point cloud analysis models, offering an innovative solution for real-time and resource-constrained applications.

\subsection{Performance Across Point Cloud Densities}
The table \ref{tab-SA-MLP-density} presents the classification accuracy of four models—Mul-MLP, Shift-MLP, Add-MLP, and SA-MLP—on the ModelNet40 dataset, evaluated with varying point cloud densities: 1024, 512, 256, and 128 points. Across all models, there is a clear trend: as the number of points decreases, the classification accuracy generally declines. This pattern is expected, as a lower number of points reduces the amount of information available for feature extraction, making it more challenging for the models to accurately classify the point clouds. However, the extent of this decline varies among the models, providing insight into their robustness to changes in point cloud density.

\begin{table}[ht]
    \centering
    \newcommand{\tabincell}[2]{\begin{tabular}{@{}#1@{}}#2\end{tabular}}
    \small
    \setlength{\tabcolsep}{5.0mm}
    \begin{tabular}{l|cccc}
        \toprule[1pt]
        Model                   & 1024  & 512   & 256   & 128   \\
        \midrule[0.3pt]
        Mul-MLP                 & 93.5  & 92.9  & 92.7  & 90.5  \\
        Shift-MLP               & 93.3  & 92.7  & 91.7  & 89.1  \\
        Add-MLP                 & 93.1  & 92.6  & 90.2  & 87.0  \\
        SA-MLP                  & 93.9  & 93.1  & 91.7  & 90.0  \\
        \bottomrule[1pt]
    \end{tabular}
    \vspace{5pt}
    \caption{Classification accuracy of Mul-MLP, Shift-MLP, Add-MLP, and SA-MLP models on ModelNet40 across varying point cloud densities.}
    \label{tab-SA-MLP-density}
\end{table}

The comparative analysis of the first three models—Mul-MLP, Shift-MLP, and Add-MLP—reveals a consistent order of accuracy across different point densities, with Mul-MLP performing the best, followed by Shift-MLP, and then Add-MLP. This trend underscores the inherent differences in the operations that underpin each model. Mul-MLP, which relies on traditional multiplication operations, provides the most stable and accurate performance across all densities due to its fine-grained feature extraction capabilities. Shift-MLP, though slightly less accurate than Mul-MLP, remains close in performance, leveraging the efficiency of bitwise shift operations. However, the slight drop in accuracy for Shift-MLP indicates that while shift operations are computationally efficient, they may lack the precision needed for capturing detailed features as effectively as multiplication-based operations.

Add-MLP exhibits the most significant performance degradation as the point cloud density decreases. This decline is particularly pronounced at lower densities (256 and 128 points), where the model’s accuracy drops more sharply compared to Mul-MLP and Shift-MLP. The sensitivity of Add-MLP to reduced point cloud density can be attributed to the nature of the adder layer, which relies on the $L_1$ norm to measure the relevance between input features and weights. The $L_1$ norm's strict requirement for matching weights and features makes Add-MLP more susceptible to noise and data sparsity, leading to less effective feature extraction when fewer points are available. As the point cloud becomes sparser, the model struggles to maintain high accuracy, reflecting the limitations of additive operations in handling lower-density data.

The SA-MLP model, which integrates both shift and add operations, demonstrates a significant performance advantage over the baseline Mul-MLP model at higher point densities (1024 and 512 points). This improvement highlights the benefits of combining the complementary strengths of shift and add operations, allowing SA-MLP to better capture a broader range of features with minimal parameter overhead. However, as the point density decreases, the performance of SA-MLP shows a slight decline compared to Mul-MLP, particularly at the lowest density of 128 points. This decrease is primarily due to the inclusion of the adder layer, which, as discussed earlier, is more sensitive to noise and data sparsity. Despite this, SA-MLP maintains competitive performance, demonstrating that the hybrid approach offers a balanced trade-off between computational efficiency and accuracy, even under challenging conditions.

When comparing SA-MLP with Shift-MLP and Add-MLP, SA-MLP consistently outperforms both across all point densities. The hybrid nature of SA-MLP enables it to leverage the advantages of both shift and add operations, making it more robust and adaptable to varying input conditions. The shift layers contribute to maintaining high computational efficiency, while the add layers enhance the model's ability to capture complex feature interactions. This combination allows SA-MLP to achieve superior accuracy and resilience, making it a particularly effective model for point cloud classification tasks where both performance and resource efficiency are crucial.

In summary, the experimental results underscore the effectiveness of the SA-MLP model in combining shift and add operations to create a robust and efficient architecture for point cloud classification. While each type of operation has its strengths and weaknesses, the hybrid approach of SA-MLP allows it to maintain high performance across different input densities, making it a promising direction for future research in the development of advanced neural network models.

\subsection{Analysis of Gradient Regularization Across Layer Types}

This subsection analyzes the gradient regularization applied across different layer types, specifically focusing on the adder layers compared to multiplication-based and shift-based layers. The analysis examines the root-mean-square (RMS) values of the gradients across various embedding and encoder layers, as shown in Tab. \ref{tab-SA-MLP-grad}. The table categorizes the different layers based on their operational types: multiplication (mul.), shift, and addition (add), with special consideration for the adder layers’ vanilla gradients before adaptive modulation (add (van.)).

\begin{table}[ht]
    \centering
    \newcommand{\tabincell}[2]{\begin{tabular}{@{}#1@{}}#2\end{tabular}}
    \setlength{\tabcolsep}{1.8mm}
    \begin{tabular}{lccccccc}
        \toprule[1pt]
                            &    
                            &\multicolumn{4}{c}{Embedding Layers}   &\multicolumn{2}{c} {Encoder Layers} \\
        \cmidrule(lr){3-6} \cmidrule(lr){7-8}
        \textbf{Model}      & type    
                            & \hspace{0.5em}\tabincell{c}{$l$-1}      & \tabincell{c}{$l$-2}
                            & \tabincell{c}{$l$-3}      & \hspace{-0.5em}\tabincell{c}{$l$-4}
                            & \hspace{1em}\tabincell{c}{$l$-1}      & \tabincell{c}{$l$-2} \\  
        \midrule[0.3pt]
        Mul-MLP             & \tabincell{c}{mul.\\($\times10^{-5}$)}    
                            & \hspace{0.5em}21.6  & 7.88  & 4.79  & \hspace{-0.5em}3.73  & \hspace{1em}8.77  & 2.58  \\
        \midrule[0.3pt]
        Shift-MLP           & \tabincell{c}{shift\\($\times10^{-5}$)}   
                            & \hspace{0.5em}2.49  & 3.30  & 1.36  & \hspace{-0.5em}1.85  & \hspace{1em}4.43  & 1.14  \\
        \midrule[0.3pt]
        \multirow{4}{*}{Add-MLP}
                            & \tabincell{c}{add (van.)\\$(\times10^{-5}$)}
                            & \hspace{0.5em}8.93  & 1.25  & 0.35  & \hspace{-0.5em}0.30  & \hspace{1em}0.20  & 0.06  \\
                            \cmidrule[0.3pt]{2-8}
                            & \tabincell{c}{add\\$(\times1.0$)}
                            & \hspace{0.5em}0.20  & 0.20  & 0.20  & \hspace{-0.5em}0.20  & \hspace{1em}0.20  & 0.20  \\
        \midrule[0.3pt]
        \multirow{6}{*}{SA-MLP}              
                            & \tabincell{c}{shift\\($\times10^{-5}$)} 
                            & \hspace{0.5em}14.4  & -     & 9.56  & \hspace{-0.5em}-     & \hspace{1em}4.84  & -     \\
                            \cmidrule[0.3pt]{2-8}
                            & \tabincell{c}{add (van.)\\$(\times10^{-5}$)}
                            & \hspace{0.5em}-     & 1.03  & -     & \hspace{-0.5em}0.91  & \hspace{1em}-     & 0.07  \\
                            \cmidrule[0.3pt]{2-8}
                            & \tabincell{c}{add\\$(\times1.0$)}
                            & \hspace{0.5em}-     & 0.20  & -     & \hspace{-0.5em}0.20  & \hspace{1em}-     & 0.20  \\ 
        \bottomrule[1pt]
    \end{tabular}
    \vspace{5pt}
    \caption{Comparison of Root-Mean-Square (RMS) values of gradients across embedding and encoder layers for Mul-MLP, Shift-MLP, Add-MLP, and SA-MLP models.}
    \label{tab-SA-MLP-grad}
\end{table}

The results reveal that the RMS values of the gradients for both the shift layers in Shift-MLP and SA-MLP, as well as the multiplication-based layers in Mul-MLP, are consistently of the same order of magnitude. This consistency suggests that shift operations can effectively replace multiplication operations without significantly altering the gradient magnitude, which is critical for maintaining stable training dynamics across these models.

However, the RMS values for the raw gradients of the adder layers in Add-MLP and SA-MLP are significantly lower and vary more across layers, particularly as the network depth increases. This variability in gradient magnitude complicates the optimization process, making it challenging to apply a uniform learning rate across the network. The differences in gradient magnitude between layers within the adder layers further emphasize the need for a more tailored approach to gradient management.

To address this issue, an adaptive modulation technique is applied to the adder layer gradients, ensuring that the RMS values after regularization are consistently set at 0.2 across all adder layers. This outcome is derived from the expression for the regularized gradient \(\tilde{g}\) outlined in the Methodology section (Eq. \ref{eq-SA-MLP-modulation}). The RMS of the regularized gradient, \(\tilde{g}_{rms}\), is calculated as follows:

\begin{equation}
\tilde{g}_{rms} = \frac{\eta}{\parallel \nabla L(W) \parallel_2 / \sqrt{n}} \times \frac{\parallel \nabla L(W) \parallel_2}{\sqrt{n}} = \eta
\end{equation}

Here, \(\eta\) is set to 0.2, effectively representing the variance of the adder layer gradient after adaptive modulation. This regularization standardizes the gradient variance across all adder layers, facilitating the use of a uniform learning rate. However, it is important to note that even after modulation, the magnitude of the gradients in the adder layers remains very different from that of the multiplication and shift layers, necessitating distinct optimization strategies. Specifically, the choice of optimizer for the adder layers diverges from that used for the multiplication and shift layers. While Adam is typically used for its adaptive gradient adjustment, this could conflict with the already stabilized gradients in the adder layers. Therefore, a simpler optimizer, such as SGD, is chosen to avoid introducing further unnecessary modulation.

In summary, the regularization of the adder layer gradients plays a crucial role in maintaining training stability and enabling effective optimization. The necessity for distinct learning rates and optimizers for the different layer types highlights the inherent differences in their operational characteristics, ensuring that the SA-MLP model can leverage the strengths of each layer type while maintaining overall performance stability.

\subsection{Visualization Analysis of Encoder Features}
\begin{figure}[htbp]
    \centering
    \includegraphics[width=1.0\linewidth]{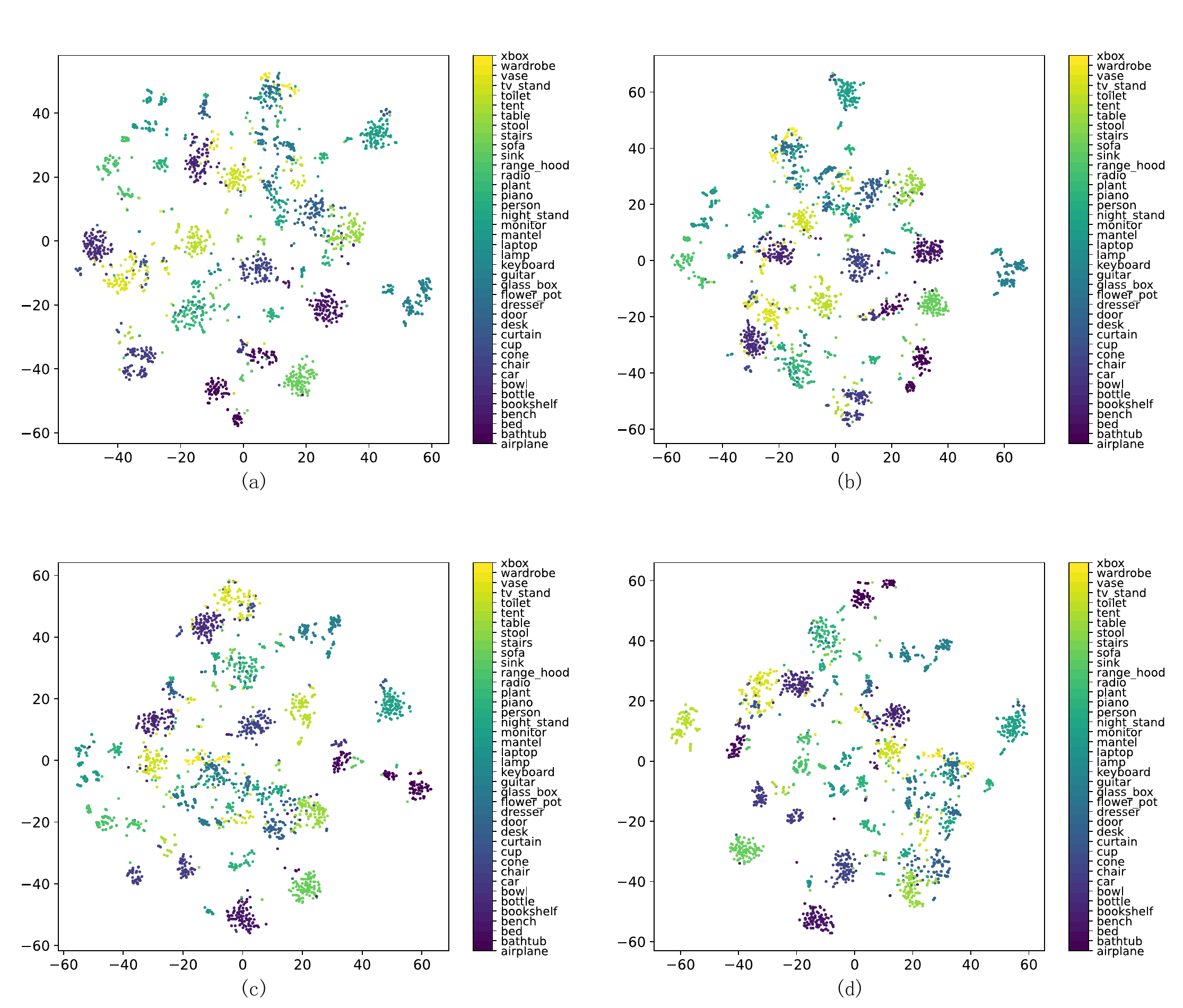}
    \caption{t-SNE visualization of the coded features from the test set for each model: (a) Mul-MLP (baseline), (b) Shift-MLP, (c) Add-MLP, and (d) SA-MLP. The visualizations highlight the feature distributions and clustering behavior of the models, with SA-MLP demonstrating tighter clustering and fewer outliers.}
    \label{fig-SA-MLP-tSNE}
\end{figure}

To further analyze the feature representations learned by each model, we applied t-distributed Stochastic Neighbor Embedding (t-SNE) visualization to the encoded features from the test set. The resulting plots are shown in Fig. \ref{fig-SA-MLP-tSNE}. The feature distribution in Fig. \ref{fig-SA-MLP-tSNE} (a) represents the baseline Mul-MLP model, which serves as the reference for comparison. Fig. \ref{fig-SA-MLP-tSNE} (a) displays a generally well-separated clustering of samples, indicating that Mul-MLP effectively captures the underlying structure of the data.

In Fig. \ref{fig-SA-MLP-tSNE} (b), which corresponds to the Shift-MLP model, the distribution is largely similar to that of the baseline Mul-MLP. This similarity suggests that replacing multiplication with shift operations does not significantly alter the model’s ability to distinguish between categories, while still offering the computational benefits of the shift operation. The clusters remain well-defined, and there is minimal overlap between different categories, indicating that Shift-MLP retains a robust feature extraction capability.

Fig. \ref{fig-SA-MLP-tSNE} (c), representing the Add-MLP model, shows some notable differences compared to the baseline. Specifically, there is an increase in overlap between categories in certain regions of the plot. This suggests that the adder layers, which rely on additive operations, may introduce some challenges in distinguishing between closely related categories. The increased repetition between categories could be due to the sensitivity of the $L_1$ norm to noise and the reduced capability of additive operations to capture complex feature relationships as effectively as multiplication.

The most significant observations arise from Fig. \ref{fig-SA-MLP-tSNE} (d), which corresponds to the SA-MLP model. Compared to the baseline, SA-MLP shows tighter clustering within categories and fewer discrete points or outliers. This improved clustering suggests that the integration of both shift and add operations enables SA-MLP to better capture the intrinsic structures of the data, leading to more compact and distinct feature representations. The fewer outliers indicate that SA-MLP may also be more resilient to variations within the data, resulting in more consistent classifications.

\subsection{Weight Distribution Analysis Across Different Layer Types}

\begin{figure}[htbp]
    \centering
    \includegraphics[width=1.0\linewidth]{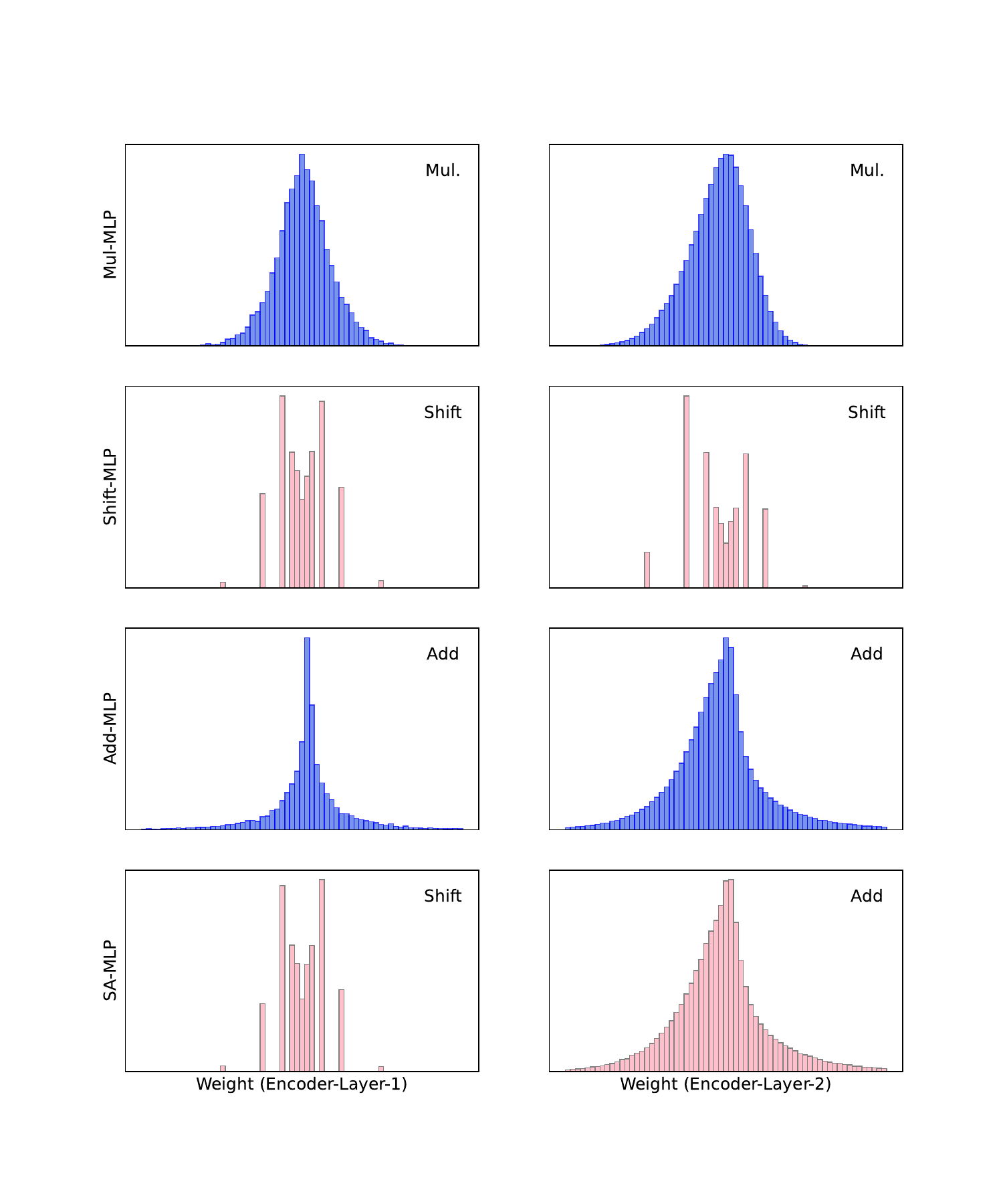}
    \caption{Histograms of weight distributions for different models and layers: (a) Mul-MLP, (b) Shift-MLP, (c) Add-MLP, and (d) SA-MLP. Each row represents a different model, while columns correspond to encoder layers. The histograms illustrate the distribution characteristics of weights across various layer types, showing differences in distribution patterns between multiplication-based, shift-based, and add-based layers.}
    \label{fig-SA-MLP-weight}
\end{figure}

In this subsection, we analyze the weight distributions of different layers across four models: Mul-MLP, Shift-MLP, Add-MLP, and SA-MLP. The analysis is conducted using statistical histograms that visualize the characteristics of the weight distributions for the two encoder layers in each model, illustrated in Fig.~\ref{fig-SA-MLP-weight}. Understanding these distributions provides insight into the underlying behavior of each model and highlights the differences between multiplication, shift, and add operations in neural networks.

In Mul-MLP, which utilizes traditional multiplication-based operations, the weight distributions for both encoder layers follow a Gaussian pattern. This pattern arises because the shared MLP weights in Mul-MLP are, under certain specific assumptions, equivalent to the $L_2$ norm, as supported by the literature~\cite{2020AdderNet}. This equivalence means that the optimization process, which minimizes the $L_2$ loss, naturally drives the weights toward a distribution centered around zero, resulting in the observed Gaussian pattern.

In contrast, Shift-MLP replaces the multiplication operations with bitwise shift operations, and as a result, the weight distribution takes on a discrete form. This is due to the quantization process inherent in shift operations, where weights are quantized to specific powers of two. The histogram for Shift-MLP reveals several discrete columns, each corresponding to a quantized value. Despite this discretization, the overall envelope of the distribution closely resembles the normal distribution observed in Mul-MLP, indicating that while the quantization introduces discrete characteristics, the general distribution pattern remains somewhat similar.

The Add-MLP model, which replaces multiplication with addition operations, shows a different pattern. The weights in Add-MLP exhibit a Laplace distribution, characterized by a sharper peak at the center and heavier tails compared to the normal distribution. This difference arises from the $L_1$ norm used in the adder layers, which penalizes large weight values more heavily than the $L_2$ norm. As a result, the optimization process in Add-MLP drives more weights to be close to zero, leading to the observed Laplace distribution. This pattern is consistent with findings in the AdderNet~\cite{2020AdderNet}, where the $L_1$ paradigm was shown to produce such distributions due to the nature of the loss function driving weights to minimize the absolute differences rather than squared differences.

Finally, the SA-MLP model, which integrates both shift and add operations, exhibits a combination of these distribution patterns. The first encoder layer in SA-MLP is a shift layer, and its weight distribution mirrors that of Shift-MLP, with discrete quantized values forming the histogram. The second encoder layer is an adder layer, and its weight distribution aligns with that of Add-MLP, showing a Laplace distribution. This combination within SA-MLP highlights the complementary strengths of both shift and add operations, as the model can leverage the efficiency of quantized shifts in the earlier layers and the fine-tuned adjustments of the adder layers in subsequent layers.

\section{Conclusion}
This study presented a series of multiplication-free MLP-based models tailored to the classification of 3D point clouds, with a primary focus on enhancing computational efficiency and reducing energy consumption, particularly in resource-constrained environments such as sensor systems and edge devices. By replacing traditional multiplication operations with shift and add operations, the proposed models, including Shift-MLP, Add-MLP, and the hybrid SA-MLP, demonstrated comparable or superior classification performance compared to conventional models. Among these, SA-MLP, which combines both shift and add operations, outperformed the baseline Mul-MLP and several state-of-the-art methods on the ModelNet40 benchmark, illustrating the potential of multiplication-free architectures as viable solutions for point cloud classification tasks, especially in sensor systems where computational resources are often constrained.

Despite these promising results, the study also identified several limitations. The training of adder layers was slower compared to traditional multiplication-based models, primarily due to the lack of optimization for addition and shift operations within current deep learning frameworks. While existing libraries such as CUDA and CuDNN are highly optimized for matrix multiplication and convolutional operations, non-standard operations like addition-based computations still lack comparable support, leading to inefficiencies during both training and inference on GPU platforms. These challenges highlight the gap in existing infrastructure for integrating multiplication-free models into widely-used deep learning workflows, thus requiring further development to fully harness the potential of these architectures in real-world applications.

To address these limitations, future research should focus on optimizing deep learning frameworks and hardware accelerators specifically for non-standard operations, such as addition and shift-based computations. Tailored optimizations for these operations could substantially improve the performance and scalability of multiplication-free architectures, making them more practical for real-time sensor data processing and enabling their broader adoption in resource-constrained environments. Furthermore, exploring advanced techniques to reduce the computational overhead of the adder layers could make these models more feasible for large-scale point cloud processing tasks, including semantic segmentation and large-scale 3D point cloud classification.

In conclusion, this work represents a significant step toward the integration of multiplication-free architectures into point cloud analysis, demonstrating that such models can substantially reduce computational consumption while maintaining high accuracy. Although the current study is preliminary, it provides a promising solution for efficient 3D point cloud classification, particularly in sensor-driven applications such as autonomous driving, robotics, and embedded systems. This work paves the way for future research to refine and optimize these models, with the potential to make a profound impact on point cloud analysis in resource-constrained environments, such as edge computing and sensor networks.

\bibliographystyle{IEEEtran}
\bibliography{reference}

\vfill

\end{document}